\documentclass[journal, 11pt, onecolumn]{IEEEtran}
\usepackage[utf8]{inputenc}
\usepackage{amsmath}
\usepackage{setspace}
\usepackage{graphicx}
\usepackage{outlines}
\usepackage{multicol}
\usepackage{hyperref}
\usepackage{todonotes}
\usepackage{xcolor}
\usepackage{geometry}
\usepackage{booktabs}
\usepackage{multirow}
\usepackage{makecell}
\usepackage[numbers,sort&compress]{natbib}
\usepackage[flushleft]{threeparttable}
\usepackage[lofdepth,lotdepth]{subfig}
\usepackage[subtle, margins=normal, bibliography=normal]{savetrees}
\pdfoutput=1

\title{\Large Integrating Psychometrics and Computing Perspectives on Bias and Fairness in Affective Computing: A Case Study of Automated Video Interviews}
\author{\small \IEEEauthorblockN{Brandon M. Booth\IEEEauthorrefmark{1}, Louis Hickman\IEEEauthorrefmark{2}, Shree Krishna Subburaj\IEEEauthorrefmark{1},\\ Louis Tay\IEEEauthorrefmark{2}, Sang Eun Woo\IEEEauthorrefmark{2}, Sidney K. D’Mello\IEEEauthorrefmark{1}}\\
\IEEEauthorblockA{Institute of Cognitive Science\IEEEauthorrefmark{1}, University of Colorado Boulder;\\ College of Health and Human Sciences\IEEEauthorrefmark{2}, Purdue University\\
Email: brandon.m.booth@gmail.com, louishickman@gmail.com, shree.subburaj@colorado.edu, stay@purdue.edu, sewoo@purdue.edu, sidney.dmello@gmail.com}}

\date{March 31$^{st}$, 2021}

\IEEEaftertitletext{\vspace{-2\baselineskip}}

\begin{document}

\IEEEtitleabstractindextext{%
\begin{abstract}
We provide a psychometric-grounded exposition of bias and fairness as applied to a typical machine learning pipeline for affective computing. We expand on an interpersonal communication framework to elucidate how to identify sources of bias that may arise in the process of inferring human emotions and other psychological constructs from observed behavior. Various methods and metrics for measuring fairness and bias are discussed along with pertinent implications within the United States legal context. We illustrate how to measure some types of bias and fairness in a case study involving automatic personality and hireability inference from multimodal data collected in video interviews for mock job applications. We encourage affective computing researchers and practitioners to encapsulate bias and fairness in their research processes and products and to consider their role, agency, and responsibility in promoting equitable and just systems.
\end{abstract}

}

\maketitle
\IEEEpubid{\begin{minipage}{\textwidth}\ \\[12pt] \centering
\copyright 2021 IEEE.  Personal use of this material is permitted.  Permission from IEEE must be obtained for all other uses, in any current or future media, including reprinting/republishing this material for advertising or promotional purposes, creating new collective works, for resale or redistribution to servers or lists, or reuse of any copyrighted component of this work in other works.
\end{minipage}}
\IEEEdisplaynontitleabstractindextext
\newpage
\section{Introduction}
The tools used in affective computing (AC), which enable machines to identify people's behaviors and mental states, are being increasingly utilized in education, healthcare, and the workplace. One application is to aid in the allocation of limited resources (e.g., counseling, mental health care, in-person interviews) via automated screening \cite{hickman2021automated, muralidhar2016training, booth2021bias}.  In these types of high-stakes scenarios, the assessments provided by AC systems can directly affect the decision processes which influence the amount of attention, care, and opportunities afforded to individuals. As such, it is important that these processes are accurate, unbiased, and fair because any deficiencies or errors present in these systems stemming from the data they were trained on, the types of algorithms used, or the decision processes themselves, may disproportionately impact different groups of people and lead to ethical and legal concerns, not to mention pain and suffering for the vulnerable groups impacted. Simply put, AC systems must deter, not propagate, extant systems of inequity and injustice.

Fortunately, we have decades of guidance on how to construct fair and unbiased measurement systems. The fields of educational and psychological measurement (i.e., \textit{psychometrics}) have well-established, distinct definitions of test \textit{bias} and \textit{fairness} \cite{american2014standards}. Great research progress is being made toward ethical data representations for artificial intelligence systems \cite{baird2020considerations} and fair emotional expression recognition systems \cite{xu2020investigating}, yet most AC research ignores psychometric aspects entirely and, when considered, many studies of algorithmic bias treat the notions of bias and fairness somewhat interchangeably (e.g., \cite{yan2020mitigating}). Thus, a crucial first step towards reducing the potential short- and long-term disparities of AC systems is forming a consistent understanding of these terms. Accordingly, one aim of this paper is to provide an exposition of bias and fairness from a psychometric perspective, to ground these terms in a typical AC machine-learning (ML) pipeline, and to enable AC researchers and practitioners to understand how sources of bias and unfairness contribute to observed manifestations or measurements of bias and unfairness.

Our contributions are as follows. First, we define the psychometric meaning of bias and distinguish it from fairness, providing examples of each. Second, we present a typical ML pipeline used in AC to generate predictions for mental constructs (e.g., emotions) from physiological and behavioral data and decompose it into a recurrent sequence of information exchanges.   We demonstrate that by representing these exchanges as noisy communication models, borrowed from classic information theory \cite{shannon1948mathematical}, one can identify possible sources of bias and unfairness at multiple stages in the pipeline. Third, we connect measures of bias and fairness from recent computer science research to the psychometric definitions of bias and fairness. Finally, using automated pre-employment screening, or \textit{personnel selection}, as an application domain which utilizes many analytical tools from AC, we empirically demonstrate the process of testing for some types of bias and unfairness in automatic personality and hireability inference from video interviews.


\section{Bias, Fairness, and Machine Learning in Affective Computing}
%
The terms \textit{bias} and \textit{fairness} are sometimes used interchangeably in reference to discrimination, and it is important to distinguish the two. Indeed, discrimination serves as an umbrella (legal) term encompassing both bias and fairness concerns \cite{CRA1964}, but these terms have distinct meanings that should not be confused.

\begin{figure}
    \centering
    \includegraphics[width=\textwidth]{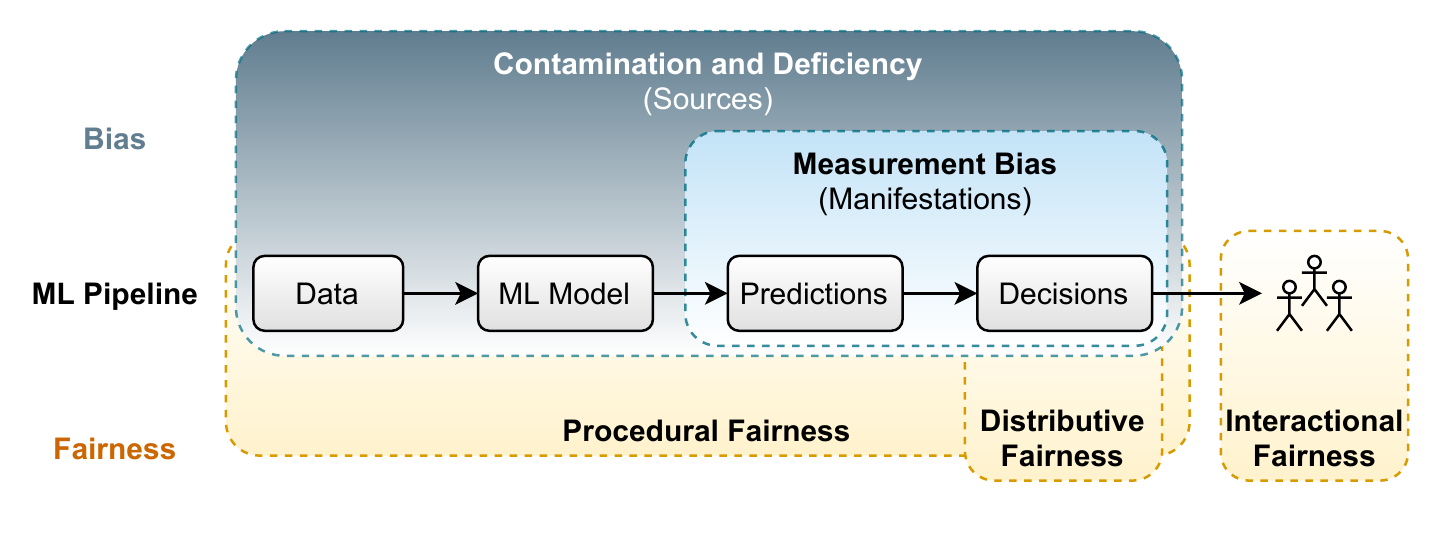}
    \caption{\small Different types of bias and fairness and their regions of concern with respect to a typical machine learning (ML) pipeline used for decision-making where the outcomes affect real people.}
    \label{fig:bias_fairness_link}
\end{figure}

The \textit{Standards for Educational and Psychological Testing} (hereafter, the \textit{Standards}) has provided guidance on the development of valid, fair, and unbiased measures since the first edition was released in 1966. In general, the \textit{Standards} provides counsel for assessments (including computational ones) of psychological constructs intended to differentiate individuals, such as for mental health treatments or educational and employment opportunities. In AC, we are often interested in measuring latent constructs (i.e.,  an individual's states or traits) which are not directly observable, such as emotion, depression, and personality.  In psychometrics (i.e., the study of psychological measurement), these constructs are measured using carefully crafted and validated assessments, including test items with correct/incorrect responses (e.g., intelligence tests), questionnaires with Likert-type scales, and other measures (e.g., observations).  In AC, these assessment items are replaced with automated inference from behaviors often obtained using cameras, microphones, and various physiological sensors. A typical ML pipeline for predicting a latent mental state involves passing data (e.g., behavioral observations about a person) through a trained ML model to obtain a prediction which can later be used to make the decisions that affect people (e.g., to hire or not hire). Figure \ref{fig:bias_fairness_link} illustrates this sequence of events and also depicts different types of bias and fairness and their regions of concern with respect to this pipeline.  


\textit{Fairness} is a subjective perspective on the appropriateness of the way a construct is measured, how the measure is used for decision-making, and explanations related to the use of the construct.  \textit{Bias} is any systematic error which differentially affects assessments of distinct groups of people. These are two different notions, but we often hear about them together because they both pertain to potential discrimination and the quality of decisions. The \textit{Standards} considers bias to be subsumed by fairness, in that a biased measure is likely to be unfair. Yet, not all measures viewed as unfair are biased, not all unbiased measures are considered fair, nor will everyone view a biased measure (and the subsequent decisions made using it) as unfair.  These terms are sometimes conflated in computer science (CS) and ML literature (e.g., \cite{mehrabi2021survey, yan2020mitigating}), but psychometrics offers a clear and established perspective on these topics.

\subsection{Fairness}
Fairness has no universal definition as it is a social, not psychometric, concept rooted in value judgments \cite{hutchinson201950}. Fairness is a subjective evaluation (e.g., justice, morality), varying across cultures and societies, and in the context of organizations such as schools, hospitals, or corporations, \textit{organizational justice}, has been the predominant theoretical concept used to recognize perceptions of unfairness \cite{cropanzano2001three}. Though AC is not broadly tied to understanding people within organizations, examining fairness through this lens is highly illustrative of some of the difficulties and inherent trade-offs (cf. \cite{kleinberg2017inherent}) in fairness considerations within AC.

Organizational justice involves three key dimensions: distributive fairness, interactional fairness, and procedural fairness \cite{cropanzano2001three}.  \textit{Distributive fairness} regards the perceived fairness of outcomes and allocations of important resources (e.g., jobs). \textit{Interactional fairness} regards how people perceive the explanations, rationales, and justifications for organizational decisions and how they perceive the interpersonal treatment they receive along the way. \textit{Procedural fairness} regards the perceived fairness of the elements of the decision-making process. Procedural fairness is emphasized in the \textit{Standards} because it is crucial that an assessment (e.g., ML predictions) does not generate different scores between subgroups if they have equivalent true scores. However, if there are differences between groups due to societal structures or biology, the assessment should accurately assess any potential differences. For example, a measure of height should not show equal heights for men and women just to be ``fair.''

Each of these types of fairness is relevant in the context of AC research, tools, and products. For example, facial recognition and expression software has been a core component of the AC toolkit and used to gain insights into the expressed emotional dynamics during social interactions.  This capability is being incorporated into ML systems that, for example, observe expression dynamics of individuals in recorded video interviews to make inferences about personality and other potentially relevant characteristics for employment \cite{hickman2021automated}. However, one well-known issue is that the underlying facial recognition software tends to be less accurate for Blacks compared to Whites \cite{buolamwini2018gender}. In this context, \textit{distributive fairness} is concerned with ways to enhance the equality of scores and outcomes for measures that include facial recognition. \textit{Interactional fairness} would be concerned with enhancing the explainability of the ML pipeline decisions and seeking to provide acceptable justifications for them. \textit{Procedural fairness} would be concerned with the use of (or error associated with) facial features for expression recognition, which may be indicative of group membership (e.g., skin color \cite{buolamwini2018gender}, face structure). 

In the United States (US), laws and case law (Title VII of the Civil Rights Act of 1964; Age Discrimination in Employment Act of 1967; Americans with Disabilities Act of 1990; Civil Rights Act of 1991; Bostock v. Clayton County Georgia) clearly define groups that are protected from employment discrimination: age, disability, race, religion or belief, sex, gender, LGBTQ status, and pregnancy or maternity.  The Civil Rights Act of 1991 established that direct or indirect measures of these group attributes \textit{cannot} be used in the decision-making process for employment.  This precedent establishes a hard line for \textit{procedural fairness} for any automated system deployed within the US and used to aid in employment decisions (other countries may have different restrictions).  By extension, this means that facial expression recognition software used to aid in employment decisions in the US cannot attempt to correct for its poorer performance for darker skin tones by being aware of skin color.  Thus, these systems must remain group-unaware (i.e., ``fairness through unawareness'') while also meeting the growing demands for fair outcomes (i.e., \textit{distributive fairness}) and explainability (i.e., \textit{interactional fairness}).  Attaining fairness in these types of systems is a difficult and complex task, both from an engineering and social perspective.

This challenge becomes even more apparent when examining recent work on ML fairness.  Many of these works emphasize \textit{distributive fairness}, which is highly desirable, but it is often difficult to achieve because of inherent trade-offs among differing perspectives on what is considered fair \cite{kleinberg2017inherent}. One perspective on distributive fairness is that of \textit{equality}, or the notion that each person or group of persons receives the same outputs (e.g., job offers). Another perspective is \textit{equity}, where each person or group of persons should receive outputs proportional to their inputs (e.g., more job offers go to those who demonstrate merit).  A third perspective is that of \textit{need}, or the notion that each person or group of persons receives outputs according to their necessity (e.g., persons lacking money or who are otherwise disadvantaged receive more job offers). These methods for distributing opportunities are fundamentally opposed, but each may be deemed fair according to individual differences in outlook.  Creating fair ML systems amounts not only to transparency and measuring fairness but also social buy-in from the organizations utilizing them and the stakeholders affected by them.

\subsection{Bias}
The term ``bias'' is semantically overloaded and has many specialized definitions in different contexts. Many are familiar with bias in the form of implicit and explicit bias, which involve systematic errors of judgment among humans due to the demographic characteristics of a given target (e.g., race, gender, religion) \cite{devine2002regulation}. These types of bias relate to systematic influences which alter human behaviors or judgments about others as a function of their group membership.

In psychological assessment, which characterizes much of the AC applications in this area, ``\textit{bias} refers to any systematic error in a test score that differentially affects the performance of different groups of test takers'' \cite[p.~23]{tippins2018principles}, where group membership is determined by distinguishing characteristics among the agents (e.g., gender, age).  For example, any facial recognition software whose accuracy scores vary by race or gender would be biased. This is the definition of bias we adopt for the remainder of this article, and first distinguish between \textit{sources} of bias and evidence or \textit{manifestations} of bias.

\begin{figure}
    \centering
    \includegraphics[width=0.4\textwidth]{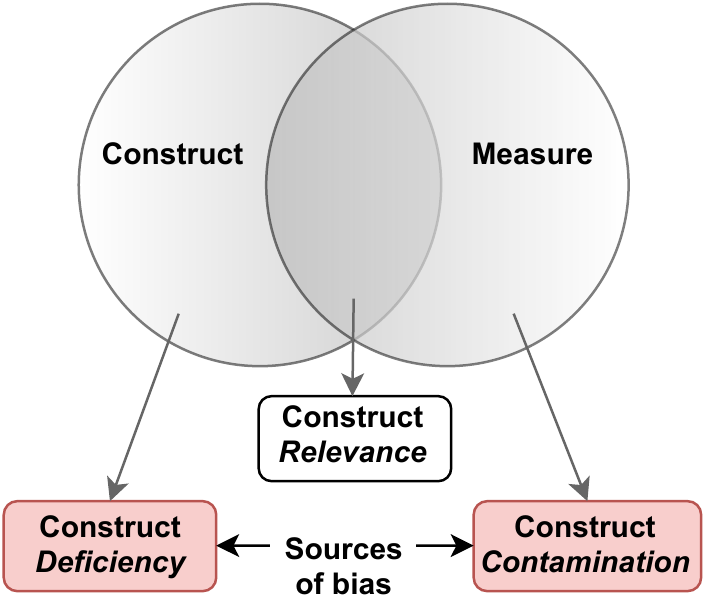}
    \caption{\small Sources of measurement bias in construct assessment \cite{american2014standards} (reproduced from \cite{booth2021bias})}
    \label{fig:bias_venn}
\end{figure}

Sources of bias in an assessment of a construct can broadly be attributed to either construct contamination or deficiency \cite{american2014standards}, as illustrated in Figure \ref{fig:bias_venn}.  \textit{Contamination} refers to sources that introduce construct-irrelevant variance, while \textit{deficiency} refers to the omission of construct-relevant variance. If these types of errors universally inflate or deflate scores independent of group status, then the assessments may be described as inaccurate, but they would not necessarily be biased.  Psychometric bias regards errors that \textit{differentially} affect members of one group compared to members of another. For example, an AC system for judging hireability from tone of voice and nonverbal behaviors (ostensibly signalling competence) trained exclusively on Whites (i.e., population bias, sampling bias, and representation bias) will tend to be deficient when assessing hireability patterns for other racial groups, and it will also be contaminated with the behavioral patterns applicable to Whites but not other racial groups.

The manifestation of bias in ML, or the bias that we observe, mostly comes from \textit{measurement bias}. Measurement bias occurs when assessment scores contain systematic error that is not relevant to the construct of interest, such as an ML pipeline producing predictions of personality scores which are systematically influenced by race. In this case, measurement bias would be observed if racial subgroups have the same ground truth scores but the assessment systematically provides different scores.

Importantly, we do not make any intentional attributions to bias, instead arguing that it arises from the involvement of humans in the process. For example, in AC, we typically rely on human-produced (i.e., self- or other-reported) assessments of constructs to serve as ground truth labels for ML modeling.  This step is necessary because the constructs of interest are latent, meaning they are hidden and cannot be directly observed.  Intuitively we may imagine that differences in the mean ground truth label for different groups indicates bias, but the \textit{Standards} states that ``group differences in outcomes do not in themselves indicate that a testing application is biased or unfair'' \cite[p.~54]{american2014standards}.  Any differences in group means may reflect true differences in the underlying construct. Simply put, a measure of height that systematically indicates that men are taller than women is not biased.

Ground truth in AC is sometimes obtained with the aid of validated self-report psychological measures, which have ostensibly been tested for bias with various subgroups.  However, when observers are used to obtain ratings or annotations, steps should be taken to ensure ground truth validity, including conducting frame-of-reference training \cite{campion1997review}, using a panel of diverse annotators, monitoring annotation quality (e.g., via inter-rater reliability/agreement), and removing outlying or low-quality annotators.  These kinds of steps result in a collection of annotations which can be better trusted in aggregate as accurate ground truth representations, where any group differences are a reflection of true differences between groups rather than bias.  These steps are absolutely essential for bias analysis--if human implicit/explicit bias contaminates the ground truth measure, then the resulting ML assessment is very likely to reflect these biases.


\section{Identifying Sources of Bias}


We endeavor to provide a framework for identifying the possible sources of bias in AC ML (Table \ref{tab:bias_types}) by deconstructing the ML pipeline into a recurrent sequence of exchanges of information between different pieces of the pipeline. We can then examine the sources of bias associated with each piece to understand how bias emerges, propagates, and manifests at various points throughout the ML process.  To do so, we expand upon a common conceptual framework employed in AC and natural language processing to understand noisy information exchange \cite{shannon1948mathematical, berlo1965process}: the two-agent communication model. We caution, however, that any list of biases, such as the examples listed in Table \ref{tab:bias_types}, only serve as a reference for researchers, and may not be comprehensive enough to represent all possible sources of bias in any given study.

\begin{table}[]
    \small
    \centering
    \caption{Sample sources of bias relevant to machine learning for affective computing}
    \label{tab:bias_types}
    \renewcommand{\arraystretch}{2.0}
    \begin{threeparttable}
    \begin{tabular}{cll}
    \toprule
& \textbf{Bias Term} & \textbf{Meaning} \\
\midrule
\parbox[t]{2mm}{\multirow{2}{*}{\rotatebox[origin=c]{90}{\textbf{Deficiency}}}}
& \makecell[l]{Selection bias /\\sampling bias} & \makecell[l]{Statistics, demographics, and user characteristics are different in\\ the user population than the collected data} \\
& Omitted variable bias & One or more important variables are left out of the model \\
\midrule
\parbox[t]{2mm}{\multirow{5}{*}{\rotatebox[origin=c]{90}{\textbf{Contamination}}}}
& Historical bias & \makecell[l]{Existing systemic biases seep into the data collection process} \\
& Representation bias & \makecell[l]{Decision makers incorrectly apply priors from an earlier\\ situation they perceive as similar to the current one} \\
& Behavioral bias & \makecell[l]{Behavior in the user population differs from behavior in the\\ training data} \\
& Presentation bias & \makecell[l]{When the order or style of information presented to participants\\ causes faulty reasoning or alters their behavior} \\
& Observer bias & \makecell[l]{Tendency for people to subconsciously project their expectations\\ onto their observations} \\
\bottomrule
    \end{tabular}
    \begin{tablenotes}
     \item \small \textbf{Note:} Although these terms are commonly dubbed \textit{bias} and may cause psychometric bias, they are not themselves \textit{psychometric bias} (i.e., systematic error differentially affecting groups).
    \end{tablenotes}
    \end{threeparttable}
\end{table}

\subsection{Communication Model for Bias Identification}
\begin{figure}
    \centering
    \includegraphics[width=\textwidth]{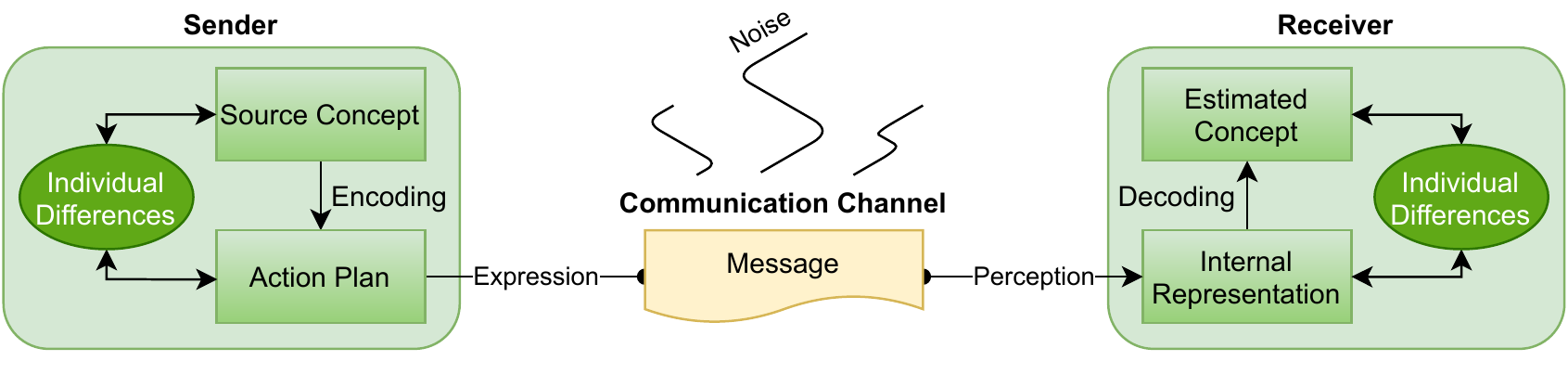}
    \caption{{\small A one-way communication model for the transmission of a source concept (information) sent by one agent (sender) and received by another (receiver).  Successful communication in this framework relies on proper encoding and decoding of a concept, moderated by each agent's individual differences, and also on minimal interference (noise) from the communication channel.  The agents' individual differences and communication channel noise represent potential sources of bias.}}
    \label{fig:one_way_comm_model}
\end{figure}

Mehu and Scherer \cite{mehu2012psycho} helped bridge social signal processing with psychology and ethology by considering how the (un)reliable and contextual nature of human behavior impacts communication and efforts to automate its understanding. In this work, they considered communication as an encoded exchange of information, hearkening back to early communication models, such as the Shannon-Weaver model \cite{shannon1948mathematical} first proposed in 1948 or the sender-message-channel-receiver (SMCR) model later introduced by David Berlo \cite{berlo1965process}. These models consider communication as a process (not necessarily serial) between a sender, who encodes a message; a receiver, who perceives and decodes the message; and a channel through which the message passes between the two, as illustrated in Figure \ref{fig:one_way_comm_model}. 



Information in this diagram flows from left to right starting with a \textit{source concept} representing what the sender intends to communicate. This concept is then encoded as a sequence of behavioral actions (the \textit{action plan}). These actions may include speaking, gesturing, touching, typing, or generally any form of sensory output. The speaker attempts to execute the actions to try to express the source concept to a receiver by means of a \textit{communication channel}, such as air (carrying vocalizations) or a digital video recording. This channel is considered to be a noisy information tunnel where the encoded message may be altered on its way to the receiver and may thus interfere with the receiver's interpretation and understanding.  The receiver perceives a potentially contaminated or deficient version of these expressions (e.g., via sensory inputs) and generates an \textit{internal representation} of the expression.  Finally, this representation is decoded to form the receiver's \textit{estimated concept}, a version of the sender's source concept.  Individual differences (e.g., experiences, behaviors, genetic traits) in the sender and receiver may separately influence their encoding and decoding of the message which, together with noise in the communication channel, are potential sources of bias.

To make this model more concrete, consider an example relevant to AC where the sender is a speaker and attempts to describe her feelings (\textit{source concept}) about a recent event to a friend. The speaker recalls her prior experiences to decide (\textit{action plan}) how to describe her feelings.  Her voice travels through the air (\textit{communication channel}) to the receiver, who perceives the utterances and forms an \textit{internal representation} of the words.  The word representations are decoded to form the receiver's \textit{estimated concept} and give meaning to the message.

In this example, the speaker's and receiver's individual differences as well as the communication channel may be sources of bias.  If the speaker has a speech impediment (\textit{speaker individual difference}) which momentarily interferes with her vocalizations, then there may be bias in the form of a deficiency in the information exchanged.  Supposing the receiver is elderly and suffers from high-frequency hearing loss, then words may be lost during perception (\textit{communication channel}).  An elderly receiver may also have more trouble remembering the entirety of the message (\textit{receiver individual differences}).  If we assess the accuracy of this information exchange based on successful transmission of the sender's source concept to the receiver, then this type of information exchange would be biased as it is systematically less accurate for people with speech impediments, hearing loss, or memory difficulties.

The model presented in Figure \ref{fig:one_way_comm_model} enables us to enumerate the potential sources of bias during information exchange: the sender's individual differences, systematic noise in the communication channel, and the receiver's individual differences. Referring back to our notions of contamination and deficiency from the \textit{Standards}, these bias sources can affect the information by contaminating it and/or facilitating omission.  We can chain elements of this model together to understand how biases influence communication in larger systems.


\subsection{Machine Learning as a Communication Process}


Let us examine a typical process for training a supervised ML model for affective computing. First, data (face, voice, physiology, actions) is collected from a group of participants, usually selected out of convenience, while they engage with a stimulus (stimuli) and complete a task (tasks). Then, for each participant, a set of labels of subjective constructs (e.g., emotions) are collected using self-annotation or a panel of human observers.  These assessments are combined or fused (e.g., by averaging) to form a ground truth representation.  Separately, a machine observer generates a set of features representing the participants' external behaviors and internal (e.g., physiological) responses.  A model is trained using the machine-observed features and the ground truth scores and subsequently tested using a predetermined evaluation metric, as defined and operationalized by stakeholders, to assess whether the model meets the goals of the project.  Each of these steps may be repeated iteratively until the model is satisfactory.

\begin{figure}
    \centering
    \makebox[\textwidth][c]{\includegraphics[width=1.0\textwidth]{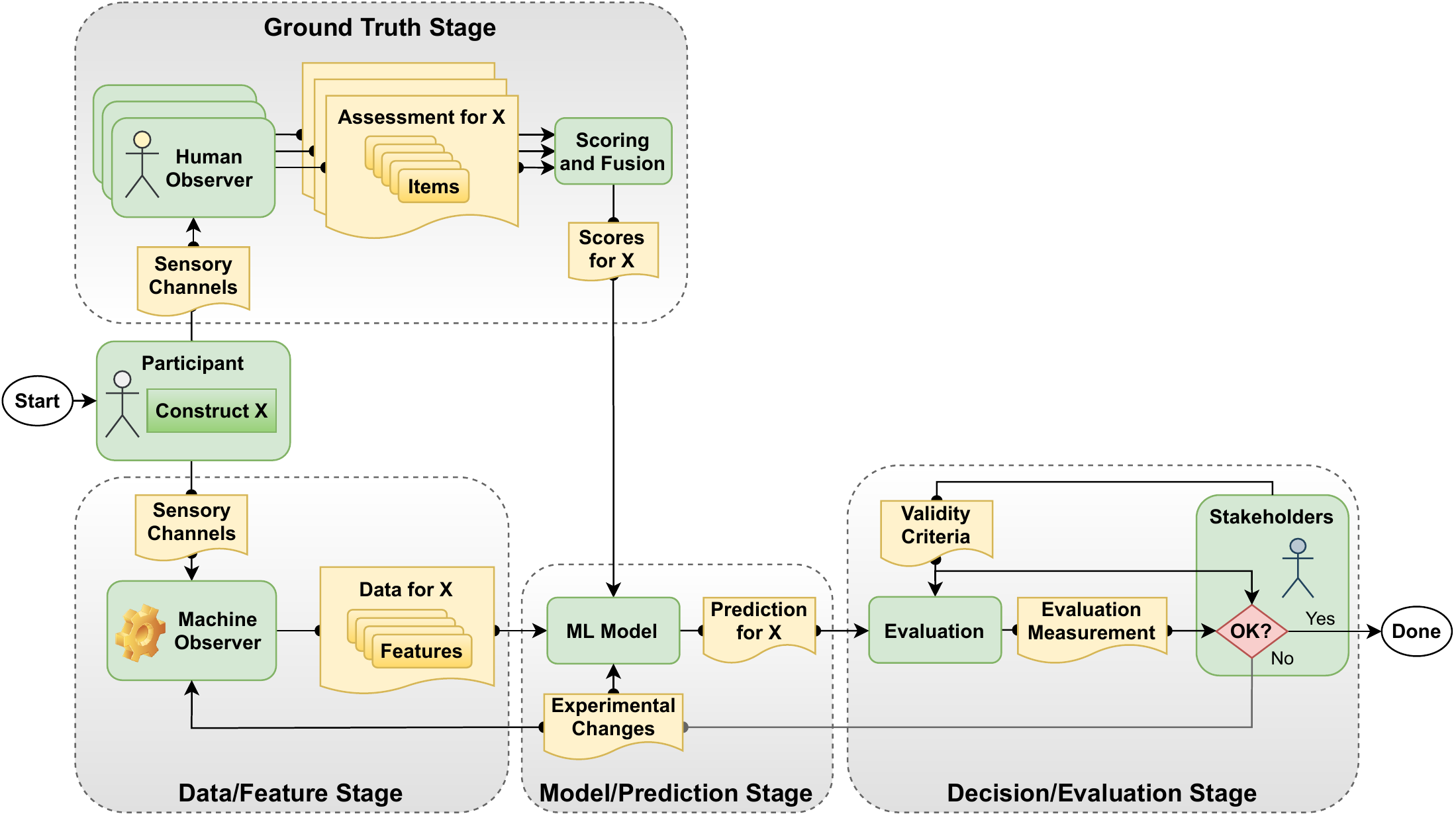}}
    \caption{{\small A deconstructed AC ML pipeline for developing a trained ML model capable of predicting a participant's construct ``X" (e.g., emotion).  Noisy information exchanges (i.e., yellow wavy boxes) between each agent (i.e., green box senders and/or receivers) in the pipeline are modeled using the noisy communication model (Figure \ref{fig:one_way_comm_model}). Elements are grouped into pipeline stages (similar to Figure \ref{fig:bias_fairness_link}) and labeled (gray dashed boxes).}}
    \label{fig:ml_communication_model}
\end{figure}

Figure \ref{fig:ml_communication_model} illustrates this process for training AC ML models, beginning with a participant (left) and ending with satisfied stakeholders (right).  This process includes several steps where information is exchanged between various agents, so we can utilize the noisy communication model from the previous section to represent the flow of information. Each information producer or consumer in the process is represented as a sender or receiver (green boxes).  The information passed between them is represented as a message channel (yellow wavy boxes).



Given this version of the ML process represented as a sequence of communication model exchanges, we can start to think about each step in the pipeline as being a potential source of bias.  Just as before, each message channel may be subjected to external noise, which may differentially corrupt or omit information transmitted from the source sender.  Likewise, the senders and receivers themselves may also introduce bias to the ML process when they send or receive messages based on their individual differences.

%

Taking the time to depict the flow of information in the ML process, whether it follows this typical AC workflow or not, enables stakeholders to produce a comprehensive set of bias sources from which construct-relevant information may be ignored (deficiency) or from which construct-irrelevant information may be added (contamination).  However, the potential impact of contamination and deficiency from each source is not equally impactful or relevant in all settings.  For example, the data produced by a machine observer is often communicated to the learning model digitally and may by temporarily stored as a file in computer memory.  The noise influencing this digital communication channel is characterized by bit corruption and/or read/write errors which are unlikely occurrences in modern robust computer hardware and even more unlikely to differentially impact groups. On the other hand, when interactions occur over videoconferencing, known deficiencies in the communication medium, such as bandwidth which vary based on socioeconomic status, can be a source of bias.

Potential bias introduced by the ML model itself, taken as another example, illustrates the key distinction between sources and manifestations of bias.  Some ML models, once fully trained, result in a fixed and deterministic algorithm which always maps the same feature inputs to a particular prediction (e.g., random forests, linear regressions, neural networks).  Once these models are deployed, they never change and thus cannot draw from prior predictions to introduce additional bias (i.e., bias other than the training data) into their predictions.  These types of models therefore \textit{cannot} be sources of bias themselves because they lack any agency and are not influenced by prior experiences (i.e., \textit{individual differences} from Figure \ref{fig:one_way_comm_model}).  Learning models which do modify future predictions based on inaccuracies in the past (e.g., online ML, reinforcement learning) have a memory and may appropriately be considered sources of bias.  When differential outcomes across groups are observed in the output from the ML model (i.e., manifested bias) and the model itself cannot be a source of bias, then the bias source is upstream.  It may be directly upstream (e.g., the machine observer), or it may stem from previous decisions made by the stakeholders (see Figure \ref{fig:ml_communication_model}) who, for example, may have selected a machine observer that is not equally accurate across groups, as in the case of facial expression recognition software that is less accurate for people with darker skin tones or eye trackers that have difficulty for those with corrective vision.

\section{Measuring Bias and Fairness}
Let us consider how to measure bias and fairness in practice. Many formalized definitions of bias and fairness have been proposed and explored in the ML literature.  We refer interested readers to \cite{mehrabi2021survey, mitchell2018prediction, verma2018fairness} for an overview. Here we focus on metrics for measuring psychometric bias and fairness when ML models are used to measure psychological constructs such as emotion or personality.

Table \ref{tab:bias_fairness_metrics} lists several bias and fairness metrics relevant to AC and categorizes them according to their inputs by pipeline stage as illustrated in Figure \ref{fig:ml_communication_model}. This list is not exhaustive and only intended to be instructive to readers when considering how bias and fairness can be measured at different ML pipeline stages.  Note that each of the three stages near the bottom of the figure (i.e., feature, prediction, and decision stages) are represented in the table, but not the ground truth stage.  This is because in AC, the constructs of interest are always latent (e.g., emotions), and thus a ground truth measure must first be established in order to evaluate bias.  Any sources of construct contamination or deficiency in the ground truth stage can be evaluated using traditional psychometric techniques (e.g., differential item functioning, differential prediction) and is not our focus here.  


\subsection{Bias Metrics}
Bias metrics come in a variety of forms and are designed to help probe for group differences (see the top half of Table \ref{tab:bias_fairness_metrics}). Each of these measures is mathematically defined, but we omit the mathematical definitions for simplicity.  Interested readers are referred to \cite{mehrabi2021survey, verma2018fairness, mitchell2018prediction} for more information with our note of caution that the terms \textit{bias} and \textit{fairness} are sometimes used interchangeably in the CS literature.

Bias metrics at the feature stage are concerned with the contamination or deficiency of construct-relevant information contained within the predictors themselves. \textit{Fairness through unawareness} is a binary metric which considers whether group membership is included as a predictor, which is often (but not always) construct-irrelevant and presumably should be excluded to minimize bias. Other measures of bias at the prediction and decision stages are tied to the accuracy of the models' predictions and are designed to check for unexpected differences in accuracy between groups. Intuitively, a prediction that is less accurate for one group compared to another contains systematic error that disproportionately affects one group over another, whether that error is introduced via contamination or deficiency. These measures can only provide evidence of manifested bias, so the bias source is always somewhere upstream of the measure (see Figure \ref{fig:ml_communication_model}).



\begin{table}[]
    \small
    \centering
    \caption{Sample bias and fairness metrics relevant to machine learning for affective computing}
    \label{tab:bias_fairness_metrics}
    \renewcommand{\arraystretch}{1.3}
    \begin{threeparttable}
    \begin{tabular}{clll}
    \cmidrule[1pt]{1-4}
    & \textbf{Stage(s)} & \textbf{Name} & \textbf{Description} \\
    \cmidrule{1-4}
    \parbox[t]{2mm}{\multirow{1}{*}{\rotatebox[origin=c]{90}{\textbf{\textbf{Bias}}}}}
    & \multirow{1}{*}{Features} & \makecell[l]{Fairness through\\ unawareness} & \makecell[l]{Group membership is not used in an assessment} \\
    \cmidrule{1-4}
    \parbox[t]{2mm}{\multirow{12}{*}{\rotatebox[origin=c]{90}{\textbf{Procedural Fairness / Measurement Bias}}}}
    & \multirow{3}{*}{Prediction} & Correlational accuracy & \makecell[l]{Equal correlations between prediction and ground\\ truth across groups} \\
    & & Differential item functioning & \makecell[l]{Equal item-total correlations for annotations\\ and/or predictions} \\
    \cmidrule{2-4}
    & \multirow{7}{*}{\makecell[l]{Prediction /\\Decision}} & Effect size difference & \makecell[l]{Effect sizes between groups in predictions are equal\\ to effect sizes between groups in ground truth}\\ 
    & & Treatment equality & \makecell[l]{Equal ratio of false negatives to\\ false positives across groups} \\
    & & Equalized odds & Equal group true and false positive rates\\
    & & Equal opportunity & Equal group true positive rates \\
    & & Predictive equality & Equal group false positive rates \\
    & & Overall accuracy equality & Equal confusion matrices across groups \\
    & & Predictive parity & \makecell[l]{Chance of individual selection across groups is\\ equal using ground truth and predictions} \\
    \midrule
    \parbox[t]{2mm}{\multirow{8}{*}{\rotatebox[origin=c]{90}{\textbf{Distributive Fairness}}}}
    & \multirow{6}{*}{\makecell[l]{Prediction /\\Decision}} & \makecell[l]{Statistical parity /\\ group fairness /\\adverse impact} & Equal group passing or hiring rates \\
    & & AUC parity & \makecell[l]{Area under the receiver operating characteristic\\ curve is equal across groups}\\
    & & Fairness through awareness & \makecell[l]{Equal predictions are given to similar\\ individuals, given group knowledge} \\
    & & Counter-factual fairness & \makecell[l]{Equal predictions are given to individuals if\\ hypothetically assigned to different groups} \\
    \cmidrule{2-4}
    & \multirow{2}{*}{Decision} & \makecell[l]{Conditional demographic\\ parity} & Decisions are independent of groups given the data \\
    & & Single-threshold & A single decision threshold is used for everyone \\
    \cmidrule[1pt]{1-4}
    \end{tabular}
    \end{threeparttable}
    \begin{minipage}{\textwidth}
    {\small The metrics are categorized according to their inputs by pipeline stage as illustrated in Figure \ref{fig:ml_communication_model}.}
    \end{minipage}
\end{table}

Bias measurements which capture group differences in accuracy depend on the measures of accuracy themselves.  For example, studies which use correlational or mean-level measures of accuracy can use group differences in these measures as relevant measures of bias. Similarly, for studies involving binary label prediction, accuracy measures derived from the confusion matrix, such as \textit{treatment equality} or \textit{equalized odds}, can be relevant bias measures.  Bias can be further evaluated at the decision stage for some decision function by examining group differences in prediction-based outcomes when compared to the outcomes resulting from applying the same decision function to the ground truth construct labels.

Once the bias metrics are computed, a separate question is how to interpret them.  When the same bias measures are reported in related research, direct comparisons can be made.  However, in the absence of comparable bias measures, differences in accuracy between groups can be difficult to interpret.  One solution is to implement multiple ML prediction experiments with small changes, perhaps involving bias reduction strategies, and then compare bias measures (with accompanying statistical tests) to assess the effects of these changes. 

\subsection{Fairness Metrics}


Fairness is not concerned with the differential group accuracy but rather how information is consumed and transformed to produce an actionable result as well as how different people are treated and impacted by the decisions. The bottom half of Table \ref{tab:bias_fairness_metrics} presents some fairness measures that are relevant in AC, but readers are referred to \cite{mehrabi2021survey} and \cite{mitchell2018prediction} for more comprehensive lists.

Each fairness measure is based on a different interpretation of fairness and is therefore not necessarily compatible with other measures of fairness.  For example, \textit{counter-factual fairness} assumes that predictions should be equal for an individual regardless of group membership, which presumes knowledge of group types and intentionally ignores true group membership for individuals.  \textit{Conditional demographic parity} suggests decisions are fair when they are independent of the relevant groups, given a particular set of data, which assumes knowledge of group types and may or may not include true group membership for individuals. \textit{Fairness through awareness} assumes that group membership contains useful information for adjusting predictions to make them more accurate and thus should be included in an assessment.  Researchers and practitioners should be aware of the underlying assumptions imposed by each metric and whether they are compatible with stakeholder goals and legal restrictions.


Additionally, fairness metrics are distinct from measurement bias metrics and need to be considered separately in order for stakeholders to evaluate the benefits and potential harms of a deployed automated AC assessment tool.  In a hypothetical experiment using real-world data, Kleinberg et al \cite{kleinberg2018algorithmic} have shown that an algorithm for deciding college admissions that is given knowledge about demographics (e.g., \textit{fairness through awareness}) could help inform admissions committees in admitting a greater proportion of Black and African American students (who are underrepresented in US colleges) while also meeting or exceeding average student body GPA goals. Notably, any algorithm that utilizes demographics to measure some construct is inherently biased because the algorithm is using construct-irrelevant information (e.g., race) in order to measure the construct.  This is one perhaps counter-intuitive example where increasing the bias of an assessment can lead to more fair outcomes.


Finally, some of the proposed measures of fairness make strong assumptions about the true distribution of construct scores in the population across groups which may or may not be reflected in the ground truth.  For example, \textit{statistical parity}, \textit{group fairness}, and \textit{adverse impact} are all concerned with the equality of acceptance rates across groups, presuming that ground truth group score distributions should be equal. The \textit{Standards} explicitly rejects these definitions of fairness because real differences may exist between groups (e.g., women tend to be perceived as slightly more extroverted than men \cite{weisberg2011gender}), however, it points out that group differences should cause additional scrutiny for other potential sources of measurement unfairness and bias. In certain high-stakes scenarios where automated AC assessment tools are used, such as employee selection in the United States, noticeably different hiring rates (\textit{adverse impact}) may constitute \textit{prima facie} evidence of discrimination \cite{CRA1964}.  In spite of the strong (and sometimes perhaps inaccurate) assumptions made by these fairness measures, it is crucial that researchers and practitioners engaged in developing high-stakes AC systems evaluate and use them to avoid any ethical or legal concerns and to avoid harming vulnerable populations.

\section{Case Study: Automated Video Interviews}


We demonstrate the process of mapping and measuring potential sources of bias and fairness in a case study of automated video interviews (AVIs).  In AVIs, job candidates are given a series of questions and asked to record their answers as part of a one-way (or asynchronous) interview.  AVIs use computer software to ingest the recordings and generate behavioral features, which are inputted to ML models to score interviewee knowledge, skills, abilities, or other characteristics (e.g., personality) to help companies screen the candidates (e.g., a yes or no decision about whether to proceed with in-person interviews or hiring; \cite{hickman2021automated}).  Human annotations are often used during the ML model development process as ground truth reference.  Human assessments of these traits are based on the dynamics of vocalization, body expression, linguistic cues, perceived emotions, and other social signals as collected by speech and natural language processing, computer vision, and various other AC tools.

Fortune 500 companies are increasingly interested in utilizing AVIs to help screen job candidates more efficiently and effectively, but there has recently been push back due to potential biases in these systems \cite{raghavan2020mitigating}.  For instance, in an in-person mock job interview experiment, Muralidhar et al \cite{muralidhar2016training} observed that the automated assessment often rated males substantially higher than females on professional, social, and communication skills (e.g., enthusiasm, competence, motivation), postulating the differences were due to gender stereotyping in social cue perception while collecting ground truth scores. It is both legally and ethically imperative that developers of these high-stakes AVI systems carefully analyze bias and fairness to avoid social harm and aid in promoting just systems.  We demonstrate this process using a data set of mock video interviews and an ML model trained to make assessments of job-relevant traits.

\subsection{Data and Models}
\textbf{Description:}  A total of 511 college students (62\% female, 37\% male, 1\% non-binary) were recruited to participate in a mock video interview for a hypothetical job opening.  Participants were presented with six interview questions in random order one at a time, and for each question they were given a few minutes to prepare a response before recording a short video (1-3 minutes) of themselves answering it.  A full list of questions can be found in \cite{booth2021bias}; one sample question is, ``tell me about a recent uncomfortable or difficult work situation. How did you approach this situation? What happened?".


\textbf{Ground truth:} At least three members of a larger panel of trained human annotators, acting as the interview committee, rated the videos for each participant.  The seven constructs of interest were Big Five personality traits (agreeableness, openness, extraversion, emotional stability, conscientiousness), perceived intellect, and hireability. The ratings were provided separately by each rater and were based on all responses from a given participant.  After establishing adequate inter-rater reliability (one-way, random, average intra-class correlation coefficient ICC[1,k] $=0.67$), the ratings from the annotators were averaged to generate a ground truth score for each participant and construct.

\textbf{Features:} A set of features was extracted from each video's visual and audio channels capturing verbal (e.g., \textit{n}-gram [word and phrases] frequencies, Linguistic Inquiry and Word Count [LIWC] categories), paraverbal (e.g., loudness, Mel-frequency cepstral coefficients [MFCCs], jitter, shimmer), and non-verbal (e.g., facial action units, total body motion) behaviors. Unigram, bigram, and trigram features were computed from the audio transcripts produced by the IBM Watson automatic speech recognition service.  Bigrams and trigrams with a point-wise mutual information (PMI) less than 4.0 were dropped to reduce the overall feature count (per \cite{park2015automatic}).  For each participant, a set of statistical functionals was independently applied to the features extracted from each of the six videos, including median, standard deviation, minimum, maximum and range, and then averaged across the recordings to produce one multidimensional observation per participant throughout the entire mock interview. 


\textbf{ML Model:} A random forest learning model was selected to make predictions of the constructs based on the audio and video features (though any model would suffice for the following bias/fairness analysis).  The data set was partitioned separately for each construct into five equally sized folds utilizing a stratified sampling approach such that each construct's ground truth distributions were roughly equal across folds.  Since each data sample corresponded to a unique participant, the folds were participant-independent.  Nested five-fold cross-validation was used to tune hyperparameters of the random forest algorithm (number of decision trees \{10, 250, 500\}, maximum depth \{10, 50\}) and the verbal feature extractor (stop words \{none, English\}, minimum term frequency \{0.01, 0.02, 0.03\}).  In total, there were approximately 7877 features after PMI filtering depending on the fold, construct, and the words uttered by participants, comprised of 250 visual, 125 paraverbal, and around 7502 verbal features.
\subsection{Bias and Fairness Results}
We evaluate the bias and fairness of the AVI ML pipeline with respect to gender at the feature stage, prediction stage, and decision stage, in line with the stages illustrated in Figure \ref{fig:ml_communication_model} and mentioned in Table \ref{tab:bias_fairness_metrics}.  Our data set contained only four participants with non-binary gender affiliations, so we exclude them in the following analysis, noting that more data would be necessary to understand how bias and fairness concerns impact the excluded gender groups.

\setlength\tabcolsep{0.2em}
\begin{table}[]
    \centering
    \begin{threeparttable}
    \caption{Example bias and fairness measures in our case study}
    \label{tab:bias_fairness_results}
    \begin{tabular}{lcccc|ccc|cc}
        \toprule
        \multirow{2}{*}{\textbf{Construct}} & \multicolumn{4}{c}{\makecell{\textbf{Spearman} $\boldsymbol \rho$}} & \multicolumn{3}{c}{\makecell{\textbf{Cohen's d}}} & \multicolumn{2}{c}{\makecell{\textbf{AI Ratio}}}\\
        & \textbf{All} & \textbf{Women} & \textbf{Men} & \textbf{Women-Men} & \textbf{True} & \textbf{Pred} & \textbf{True-Pred} & \textbf{True} & \textbf{Pred}\\
        \midrule
        \textbf{Agreeableness} & .03 & .01 & .08 & -.07 & -.13 & -.22 & .09 & \textbf{.77} & \textbf{.48} \\
        \textbf{Openness} & .34 & .37 & .27 & .10 & -.13 & -.39 & \textbf{.26} & .92 & \textbf{.32} \\
        \textbf{Emotional stability} & .31 & .28 & .21 & .07 & .36 & .66 & \textbf{-.30} & \textbf{.57} & \textbf{.31} \\
        \textbf{Conscientiousness} & .33 & .34 & .23 & \textbf{.11} & -.34 & -.61 & \textbf{.27} & \textbf{.36} & \textbf{.14} \\
        \textbf{Extraversion} & .47 & .42 & .54 & \textbf{-.12} & -.09 & -.49 & \textbf{.40} & .84 & \textbf{.36} \\
        \textbf{Perceived intelligence} & .40 & .39 & .43 & -.04 & -.06 & -.29 & \textbf{.23} & \textbf{.64} & \textbf{.64} \\
        \textbf{Hireability} & .43 & .43 & .44 & -.01 & -.11 & -.37 & \textbf{.26} & 1.0 & \textbf{.70} \\
        \bottomrule
    \end{tabular}
    \end{threeparttable}
    \begin{minipage}{\textwidth}
    \vspace{0.3em}
     \small
     \textit{Correlational accuracy} (Spearman $\rho$), \textit{effect size difference} (Cohen's d), and \textit{adverse impact} (AI) measures of bias and fairness computed in our AVI case study.  Spearman's ``Women-Men" column shows the difference in accuracy across genders.  Cohen's d ``True-Pred" column shows the difference in effect size between women and men in the true vs. predicted construct labels.  AI ratios are computed separately on the true and predicted labels.  Bold Spearman and Cohen's d values denote small effect sizes ($|\rho| > 0.1$ , $|$d$| > 0.2$), which should arouse suspicion and warrant further investigation. Bold adverse impact ratios fall below the 80\% threshold and would be \textit{prima facie} evidence of adverse impact by the ``four-fifths" rule. True = ground truth, Pred = ML model predictions.
    \end{minipage}
\end{table}

\textbf{Feature Stage:} We adopt a \textit{fairness through unawareness} strategy to minimize bias, where the gender of each participant is not included in the features used to train the ML model.  By this definition, personality, intelligence, and hireability do not depend on gender information, so including gender would contaminate the ML predictions with construct-irrelevant information and introduce bias.  While omitting gender seems to satisfy this bias goal, we note that gender information is often encoded in other features such as vocal pitch, shimmer, and MFCCs, which we do include.  These features likely contain both construct-relevant information and gender bias, so careful analysis of the impact of bias would be necessary.

Exploration of bias mitigation strategies is outside the scope of this paper, but various techniques such as the exclusion of gender-biased features (i.e., features that carry a lot of gender-relevant information) \cite{booth2021bias} or fair representation learning should be considered and tested.  Though per-gender normalization of features may seem justifiable, the US Civil Rights Act of 1991 explicitly outlaws using demographic information to indirectly adjust scores.

\textbf{Prediction Stage:} We evaluate the ML model using Spearman correlation, which examines the rank order consistency between predicted and ground truth scores, a relevant metric when applicants will be ranked against each other \cite{stachl2020personality}. The left section of Table \ref{tab:bias_fairness_results} shows correlations for all participants, separately for women and men, and also the correlation difference between women and men, which provides one measure of bias.  By themselves, these measures are difficult to interpret but would serve as a baseline for comparison in attempts to mitigate bias.  Larger group correlational differences such as the -.12 for extraversion or .11 for conscientiousness arouse our suspicions as evidence of manifested gender bias and warrant further investigation.

The second section of Table \ref{tab:bias_fairness_results} shows an \textit{effect size difference} bias measure operationalized using Cohen's d effect size $\Big($Cohen's d$ = \frac{\bar{x}_m - \bar{x}_w}{s}$, where $s^2 = \frac{(n_m-1)s^2_m + (n_w-1)s^2_w}{n_m + n_w - 2}$, $m=$men, $w=$women$\Big)$.  Mean effect size differences between men and women are computed separately in the ground truth labels and in the predictions, and then the difference in Cohen's d between the two constitutes a measure of bias. General guidelines for interpreting $d$ dictate that values between .2-.5 are considered small to medium effect sizes, and thus we can see evidence of potential bias in the predicted values for all constructs except agreeableness (for which predictions were so inaccurate that invalidity is a bigger concern than bias). A further investigation reveals that the predicted values from the ML model range between approximately 3.0-6.0 while the ground truth ranges from 1.0-7.0.  The restricted range of the predictions may be contributing to the larger differences in $d$ (due to lower standard deviation) and warrant further investigation.

\textbf{Decision Stage:} To assess compliance with the US Civil Rights Act of 1964 \cite{CRA1964}, we focus on \textit{adverse impact} (AI) as a decision-stage fairness measure.  AI is defined as the quotient of group selection ratios, in our case computed by $\min\Big(\frac{SR_W}{SR_M}, \frac{SR_M}{SR_W}\Big)$, where $SR_W$ is the number of women accepted by some binary decision process divided by the total women applicants (likewise for men).  We simulate a realistic decision function by selecting the top $k$ candidates among all participants based on the construct predictions, and then we set $k$ equal to 10\% of all participants so that we have a sufficiently large sample size for computing group selection ratios.

Taking hireability as an example trait and using the ground truth scores as a baseline, we find the selection ratio for women is .10 and for men is .10, resulting in an AI ratio = 1.0 which suggests that selecting for hireability in the ground truth is equitable.  Using hireability predictions, the selection ratio for women is .11 and for men is .08, yielding an adverse impact ratio of .70. In the US, this violates the ``four-fifths rule'' (29 CFR\S1607.4) and would be considered \textit{prima facie} evidence to support a legal discrimination claim.  Although the underlying cause of this unfairness may be bias upstream from the selection procedure, any system deployed for employment selection (at least in the US) needs to demonstrate compliance with the four-fifths rule regardless of whether any bias can be found. Further exploration of the (un)fairness of other decision thresholds (i.e., other settings for $k$) should be conducted to assess the sensitivity of the adverse impact ratio for this AVI system.

\section{Discussion} \label{sec:discussion}

This tutorial has presented a framework for understanding psychometric bias and fairness according to the \textit{Standards} in the context of machine-based assessment of emotion and related constructs.  We aimed to demonstrate that deconstructing a complex ML pipeline into a recurrent sequence of information exchanges and then treating those exchanges as noisy communication channels facilitates understanding how bias emerges, propagates, and manifests at various point throughout the ML development process.  We suggest that decomposing other complex systems involving information exchange in this same manner will enable more prescriptive bias and fairness assessments.

We were not able to cover many important issues in this tutorial and want to conclude with some remarks about future considerations.  It has been recognized that there is often a trade-off between creating the most accurate model possible and reducing bias or enhancing fairness \cite{mehrabi2021survey}. Model validity, bias, and fairness are all crucial considerations for automated AC systems, and therefore ongoing and future work should consider holistic optimization approaches rather than more traditional optimization of accuracy alone.  More work is needed to investigate and normalize methods that effectively maximize these outcomes, but some Pareto optimization techniques already show promise \cite{song2017diversity}.

In summary, researchers and practitioners in AC developing the algorithms, software, and tools employed to aid in decision making have an ethical and moral responsibility to assess systemic errors and gauge the disproportionate impact that they can impose on people.  We hope this exposition has been instructive in understanding, identifying, and measuring bias and unfairness, taking the first step toward this goal.


\section*{Acknowledgments}
This research was supported by the National Science Foundation (IIS 1921087 and IIS 1921111) and the NSF National AI Institute for Student-AI Teaming (iSAT) (DRL 2019805). The opinions expressed are those of the authors and do not represent views of the NSF.




\begingroup
\setstretch{1.03}
\bibliographystyle{IEEEtran}
\bibliography{IEEEabrv,references}
\endgroup

\end{document}